\pdfoutput=1

\documentclass{article}

\newcommand{\eat}[1]{}

\usepackage{latexsym}
\usepackage{amsfonts}
\usepackage{amsmath}
\usepackage{xcolor}
\usepackage{colortbl}
\usepackage{epsfig}
\usepackage{xspace}
\usepackage{graphicx}
\usepackage{subfigure}
\usepackage{paralist}
\usepackage{enumerate}
\usepackage{bm}
\usepackage{enumitem}
\usepackage[all]{xy}
\usepackage{cite}
\usepackage{comment}
\usepackage{booktabs}
\usepackage{balance}
\usepackage{stmaryrd}
\usepackage{pifont}
\usepackage{hhline}
\usepackage{mathabx}
\usepackage{listings}
\usepackage{array}
\usepackage{float}
\usepackage[flushleft]{threeparttable}

\usepackage[utf8]{inputenc} %
\usepackage[T1]{fontenc}    %

\usepackage{url}            %
\usepackage{booktabs}       %
\usepackage{amsfonts}       %
\usepackage{nicefrac}       %
\usepackage{microtype}      %
\usepackage{xcolor}         %

\usepackage{mathrsfs}
\usepackage{makecell}
\usepackage{xparse}
\usepackage{wrapfig}

\usepackage{epsfig}
\usepackage{multirow}
\usepackage{xurl}

\usepackage{graphicx}

\newcommand{\stab}{\vspace{1.2ex}\noindent}

\newcommand{\bi}{\begin{itemize}}
\newcommand{\ei}{\end{itemize}}

\newcommand{\be}{\begin{enumerate}}
\newcommand{\ee}{\end{enumerate}}
\newcommand{\beqn}{\begin{eqnarray*}}
\newcommand{\eeqn}{\end{eqnarray*}}

\newcommand{\stitle}[1]{\stab\noindent{\bf #1}}

\newcommand{\xmark}{\ding{55}}%
\newcommand{\sys}{CRAG\xspace}%

\NewDocumentCommand{\nan}{ mO{} }{\textcolor{blue}{\textsuperscript{\textit{Nan}}\textsf{\textbf{\small[#1]}}}}

\usepackage[final,nonatbib]{neurips_data_2024}
\usepackage{xcolor}        
\usepackage{authblk}
\usepackage[hidelinks]{hyperref}

\newcommand*\samethanks[1][\value{footnote}]{\footnotemark[#1]}

\title{CRAG -- Comprehensive RAG Benchmark}

\author[1]{Xiao Yang\thanks{Equal contribution. Correspondence to: Xiao Yang (xiaoyangfb@meta.com).}~~}
\author[1]{Kai Sun\samethanks~~}
\author[3]{Hao Xin\samethanks~~}
\author[3]{Yushi Sun\samethanks~~}
\author[1]{Nikita Bhalla}
\author[4]{Xiangsen Chen}
\author[1]{Sajal Choudhary}
\author[1]{Rongze Daniel Gui}
\author[1]{Ziran Will Jiang}
\author[4]{Ziyu Jiang}
\author[1]{Lingkun Kong}
\author[1]{Brian Moran}
\author[1]{Jiaqi Wang}
\author[1]{Yifan Ethan Xu}
\author[1]{An Yan}
\author[4]{Chenyu Yang}
\author[1]{Eting Yuan}
\author[1]{Hanwen Zha}
\author[3,4]{Nan Tang}
\author[3,4]{Lei Chen}
\author[1]{Nicolas Scheffer}
\author[1]{Yue Liu}
\author[1]{Nirav Shah}
\author[1]{Rakesh Wanga}
\author[1]{Anuj Kumar}
\author[2]{Wen-tau Yih}
\author[1]{Xin Luna Dong}
\affil[1]{Meta Reality Labs, $^\text{2}$ FAIR, Meta, $^\text{3}$ HKUST, $^\text{4}$ HKUST (GZ)}

\begin{document}
\maketitle

\newcommand{\yushi}[1]{\textcolor{black}{#1}}
\newcommand{\revision}[1]{\textcolor{black}{#1}}

\begin{abstract}
Retrieval-Augmented Generation (RAG) has recently emerged as a promising solution to alleviate Large Language Model (LLM)’s deficiency in lack of knowledge. Existing RAG datasets, however, do not adequately represent the diverse and dynamic nature of real-world Question Answering (QA) tasks. To bridge this gap, we introduce the \textbf{Comprehensive RAG Benchmark (\sys)}, a factual question answering benchmark of 4,409 question-answer pairs and mock APIs to simulate web and Knowledge Graph (KG) search. CRAG is designed to encapsulate a diverse array of questions across five domains and eight question categories, reflecting varied entity popularity from popular to long-tail, and temporal dynamisms ranging from years to seconds.  
Our evaluation of this benchmark highlights the gap to fully trustworthy QA. Whereas most advanced LLMs achieve $\le 34\%$ accuracy on CRAG, adding RAG in a straightforward manner improves the accuracy only to 44\%. State-of-the-art industry RAG solutions only answer $63\%$ of questions without any hallucination.
CRAG also reveals much lower accuracy in answering questions regarding facts with higher dynamism, lower popularity, or higher complexity, suggesting future research directions.
The \sys benchmark laid the groundwork for a KDD Cup 2024 challenge and attracted thousands of participants and submissions. We commit to maintaining \sys to serve research communities in advancing RAG solutions and general QA solutions. \revision{\sys is available at \url{https://github.com/facebookresearch/CRAG/}.}%

\end{abstract}

\section{Introduction}
\label{sec:introduction}
Large Language Models (LLMs) have transformed the landscape of Natural Language Processing (NLP) tasks, especially in Question Answering (QA)~\cite{yasunaga-etal-2021-qa,liu2023pre,lievin2024can,statsqa}.
Despite the advancements, the issue of hallucination persists as a significant challenge; LLMs may generate answers that lack factual accuracy or grounding~\cite{rawte2023troubling, 10.1145/3571730,sun2024large,verifai}. Studies have shown that GPT-4's accuracy in answering questions referring to slow-changing or fast-changing facts is below 15\%~\cite{vu2023freshllms}; even for stable (never-changing) facts, GPT-4's accuracy in answering questions referring to torso-to-tail (less popular) entities is below 35\%~\cite{sun2023head}. Overcoming hallucinations thus becomes a priority in building reliable QA systems~\cite{10.1145/3571730, huang2023survey}. 

\begin{figure}[t!]
  \centering
  \includegraphics[width=0.82\linewidth]{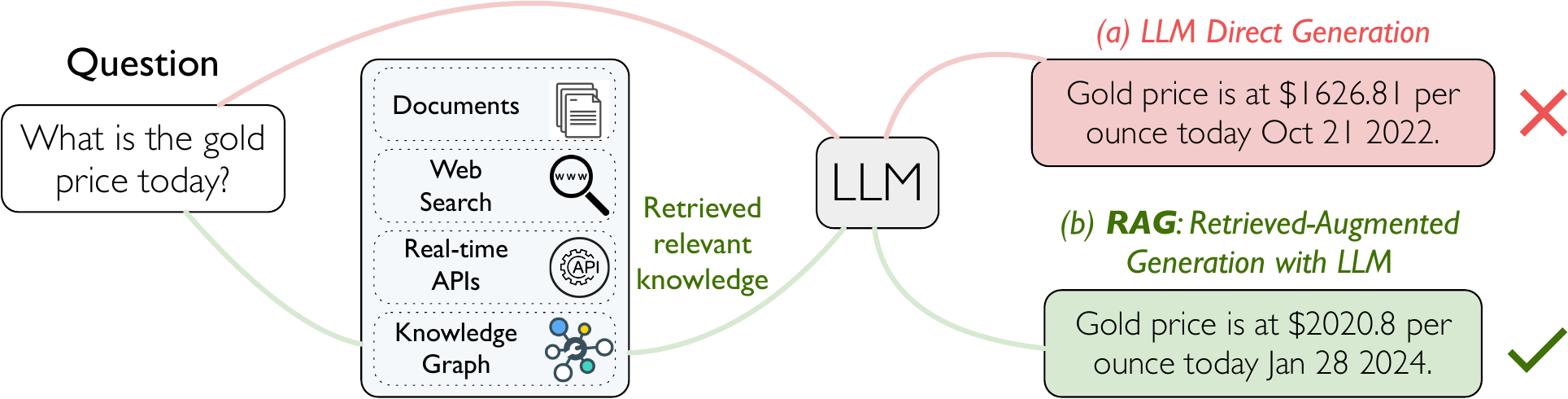}
  \caption{QA using LLMs (a) without RAG vs. (b) with RAG.}
  \label{fig:RAG}
\end{figure}

{\em Retrieval-Augmented Generation (RAG)}~\cite{gao2024ragsurvey,lewis2021retrievalaugmented,chen2023benchmarking,symphony} has recently emerged as a promising solution to alleviate LLM’s deficiency in lack of knowledge and attracted a lot of attention from both academia research and industry. Given a question, a RAG system searches external sources to retrieve relevant information and then provides grounded answers~\cite{gao2024ragsurvey, lewis2021retrievalaugmented,chen-etal-2022-murag} (see Figure~\ref{fig:RAG} for an illustration). Despite its potential, RAG still faces many challenges, such as selecting the most relevant information, reducing question answering latency, and synthesizing information to answer complex questions. %

A comprehensive benchmark is currently missing to advance continued research efforts in this field. 
Our goal is to build a benchmark that can provide a holistic view of the important capabilities and fast but reliable evaluation for RAG to propel the area forward. What is a good benchmark for QA over LLMs? We consider five critical features.

\begin{enumerate}
    \item {\bf Realism:} First and foremost, a good benchmark shall best reflect real use cases. In other words, a solution that achieves high metrics in the benchmark shall also perform very well in real scenarios. For example, the questions in a RAG benchmark shall be similar to questions people ask in real-world QA scenarios.
    \item {\bf Richness:} The benchmark shall contain a diverse set of instance types, covering both common use cases and some complex and advanced use cases, to represent real-world challenges and reveal possible limitations of existing solutions. 
    \item {\bf Insightfulness:} The benchmark shall allow for an easy understanding of performance on different slices of the data, reflecting the capability of the solution in addressing different types of challenges. 
    \item {\bf Reliability:} The benchmark shall allow reliable assessment of metrics: the ground truths shall be accurate; the metrics shall well capture the performance of the model; the evaluation shall be easy and reliable, and the computed metrics shall hold statistical significance.   
    \item {\bf Longevity:} Finally, to enable research and experimental comparison in a long term, the scenarios and the data in the benchmark shall not quickly expire and ideally shall be refreshed and improved over time.
\end{enumerate}

We strive to create a benchmark that have all of the aforementioned features, and we call it {\em CRAG -- Comprehensive benchmark for RAG}. Our work makes three contributions.

Our first contribution is the dataset itself (Section~\ref{sec:dataset}). CRAG contains a \textit{rich} set of 4,409 QA pairs from five domains: {\em Finance, Sports, Music, Movie,} and {\em Open domain}. %
In addition to simple-fact questions (asking for an attribute of an entity), CRAG contains seven types of complex questions to cover real user queries: questions with {\em Conditions}, {\em Comparison} questions, {\em Aggregation} questions, {\em Multi-hop} questions, {\em Set queries}, {\em Post-processing-heavy} questions, and {\em False-premise} questions. CRAG reflects varied entity popularity from popular to long-tail and temporal spans ranging from seconds to years, allowing easy deep dives for \textit{insights}. As we generated the questions, we referred to smart assistant use cases to make sure the questions are \textit{realistic}, paraphrased the questions to increase the \textit{diversity} of expressions, and manually verified ground truths to ensure \textit{reliability}.

In addition to QA pairs, CRAG provides mock APIs to simulate retrieval from a diverse range of available information.
This includes up to 50 full HTML pages for each question returned from a real-world search engine---the Brave Search API~\cite{brave24}, and mock KGs with 2.6 million entities. For the mock KGs, we deliberately make sure that the retrieval candidates reflect noises in a {\em realistic} setting.

Our second contribution is the evaluation mechanism to allow for {\em reliable} comparisons. We designed 3 tasks to test different components in RAG solutions: \yushi{retrieval summarization, knowledge graph and web retrieval, and end-to-end retrieval-augmented generation} (Section~\ref{sec:tasks}). Instead of computing the percentage of correctly answered questions, our score system distinguishes hallucinated answers and missing answers, and gives the former a higher penalty as it can be more harmful to ruin user trust. We also design an effective automatic evaluation mechanism to allow for fast evaluations and iterations (Section~\ref{sec:kdd_cup}).

Our third contribution is a comprehensive evaluation of straightforward RAG solutions and industry state-of-the-art solutions on RAG (Section~\ref{sec:benchmarking}). Whereas most advanced LLMs achieve $\le 34\%$ accuracy on CRAG, adding RAG in a straightforward manner improves the accuracy only to 44\%. State-of-the-art industry RAG solutions answer only $63\%$ questions without any hallucination, still having much lower accuracy in answering questions regarding facts with higher dynamism, lower popularity, or higher complexity. These evaluations serve two roles: first, they demonstrate that CRAG has appropriate level of difficulty and allows insights drawn from different dimensions of diversities the benchmark has incorporated; second, they highlight the gaps and research directions to a fully trustworthy QA system.

The \sys benchmark laid the groundwork for a KDD Cup 2024 challenge\footnote{\url{https://www.aicrowd.com/challenges/meta-comprehensive-rag-benchmark-kdd-cup-2024}}, has attracted thousands of participants and submissions within the first 50 days of the competition. We commit to maintaining \sys to serve research communities in advancing RAG solutions and general QA solutions. %

{\bf Comparison with existing benchmarks.}
Table~\ref{tab:benchmark_comparison} compares \sys with existing benchmarks for factual question answering. 
Traditional QA benchmarks such as Natural Questions (NQ)~\cite{kwiatkowski2019natural}, TriviaQA~\cite{joshi-etal-2017-triviaqa}, MS MARCO~\cite{bajaj2018ms}, and QALD-10~\cite{usbeck2023qald} have advanced QA in the past decade but \revision{consider only web retrieved \textit{or} KG retrieved contents, and} do {\em not} adequately represent the diverse and dynamic challenges that RAG is facing. New benchmarks for LLM or RAG usually target certain capabilities of the QA system. \revision{Researchers created benchmarks to evaluate how well the systems can answer simple knowledge questions~\cite{sun2023head,sun2024large,mallen2023} and handle more advanced scenarios. These include answering questions with changing answers~\cite{vu2023freshllms}, integrating information from multiple documents~\cite{chen2023benchmarking}, addressing multi-hop questions~\cite{tang2024multihop}, and answering questions with long texts~\cite{pradeep2024ragnar}. Moreover, traditional QA benchmarks usually adopt matching-based metrics such as ROUGE~\cite{lin2004rouge} or F1 to evaluate the quality of the responses~\cite{kwiatkowski2019natural,joshi-etal-2017-triviaqa}. These metrics, although working well for extractive methods, are known to not perform very effectively for LLMs that generate free-form responses~\cite{gao2024llm}.}

\revision{Despite \sys being smaller than MS MARCO and NQ}, it offers several distinctive advantages:
comprehensive coverage, realistic testing with mock APIs, dynamic question handling, diverse fact popularity, extensive content beyond Wikipedia, \revision{and fast yet reliable evaluation.}
These features make \sys a robust and versatile benchmark for testing RAG systems and broadly QA systems, providing a shared testbed to evaluate how these systems handle real-world, dynamic, and diverse information retrieval and synthesis challenges for reliable LLM-based question answering.

\begin{table*}[t]
\small
  \caption{Comparing CRAG to existing benchmarks for factual question answering.\label{tab:benchmark_comparison}}
  \centering
  \begin{tabular}{lccccccc}
    \toprule
    Benchmark & \makecell{Web \\ retrieval} & \makecell{KG \\ search} & \makecell{Mock \\ API} & \makecell{Dynamic \\ question} & \makecell{Torso and \\ tail facts} & \makecell{Beyond \\ Wikipedia} & \makecell{Question \\ size} \\
    \midrule
    QALD-10~\cite{usbeck2023qald} & \color{red}\xmark & \color{green}\checkmark & \color{red}\xmark & \color{red}\xmark & \color{red}\xmark & \color{red}\xmark & 0.8K  \\
    MS MARCO~\cite{bajaj2018ms} & \color{green}\checkmark & \color{red}\xmark & \color{red}\xmark  & not explicitly & not explicitly & \color{green}\checkmark & 100K  \\
    NQ~\cite{kwiatkowski2019natural} & \color{green}\checkmark & \color{red}\xmark & \color{red}\xmark & not explicitly & not explicitly & \color{red}\xmark & 323K \\
    RGB~\cite{chen2023benchmarking} & \color{green}\checkmark & \color{red}\xmark & \color{red}\xmark & \color{red}\xmark & \color{red}\xmark & \color{green}\checkmark & 1K \\
    FreshLLM~\cite{vu2023freshllms} & \color{red}\xmark & \color{red}\xmark & \color{red}\xmark & \color{green}\checkmark & \color{red}\xmark & \color{green}\checkmark & 0.6K \\
    CRAG & \color{green}\checkmark & \color{green}\checkmark & \color{green}\checkmark & \color{green}\checkmark & \color{green}\checkmark & \color{green}\checkmark  & 4.4K\\
    \bottomrule
  \end{tabular}
\end{table*}

\smallskip
\noindent

\section{Problem Description}
\label{sec:tasks}
A RAG QA system takes a question $Q$ as input and outputs an answer $A$; the answer is generated by LLMs according to information retrieved from external sources or directly from the knowledge internalized in the model. The answer should provide useful information to answer the question without adding any hallucination.

We designed three tasks. They share the same set of (question, answer) pairs but differ in the external data accessible for retrieval to augment QA. 
Here, we provide the content that can be leveraged in QA to ensure fair comparisons. We describe how we generated the data in Section~\ref{sec:dataset}.

\noindent
\textbf{Task 1: Retrieval Summarization.} In Task 1, we provide up to five web pages for each question. These web pages are likely, but not guaranteed, to be relevant to the question. This task aims to test the answer generation capability of a RAG system.

\noindent
\textbf{Task 2: KG and Web Retrieval Augmentation.} In Task 2, we in addition provide {\em mock APIs} to access information from underlying {\em mock KGs}. The mock KGs store structured data relevant to the questions; answers to the questions may or may not exist in the mock KGs. The mock APIs take input parameters, oftentimes parsed from the question, and provide structured data from the mocked KGs to support answer generation. This task tests how well a RAG system 1) queries structured data sources and 2) synthesizes information from different sources. \includecomment{Luna: Will we provide the same piece of information from different sources? Xiao: Luna, do you mean whether the mock APIs will have the same set of input params?}

\noindent
\textbf{Task 3: End-to-end RAG.} Similar to Task 2, Task 3 also provides both web search results and mock APIs as candidates for retrieval but provides $50$ web pages, instead of $5$, as candidates. The larger set of web pages are more likely to provide necessary information to answer the question, but meanwhile are more likely to contain noises. As such, Task 3 in addition tests how a RAG system ranks a larger number of retrieval results. 

The three tasks, each adding upon the previous one, allow testing different capabilities of the RAG systems. \revision{The only component of a RAG system not covered by these tasks is search retrieval. One may easily extend the tasks to use all 220K webpages in our benchmark as the search corpus for fully end-to-end testing.}

\section{Dataset Description}
\label{sec:dataset}

\begin{table}[t!]
\centering
\small
\caption{Definition of CRAG question types.}
\label{tab:question_type}
\begin{tabular}{p{2.6cm}p{10.5cm}}
\toprule
\textbf{Question type} & \textbf{Definition}  \\
\midrule
Simple                 & Questions asking for simple facts that are unlikely to change overtime, such as the birth date of a person and the authors of a book.  \\
\midrule
Simple w. Condition    & Questions asking for simple facts with some given conditions, such as stock prices on a certain date and a director's recent movies in a certain genre. \\
\midrule
Set                    & Questions that expect a set of entities or objects as the answer (e.g., ``\textit{what are the continents in the southern hemisphere?}''). \\
\midrule
Comparison             & Questions that compare two entities (e.g., ``\textit{who started performing earlier, Adele or Ed Sheeran?}'').\\
\midrule
Aggregation            & Questions that require aggregation of retrieval results to answer (e.g., ``\textit{how many Oscar awards did Meryl Streep win?}''). \\
\midrule
Multi-hop              & Questions that require chaining multiple pieces of information to compose the answer (e.g., ``\textit{who acted in Ang Lee's latest movie?}''). \\
\midrule
Post-processing-heavy       & Questions that need reasoning or processing of the retrieved information to obtain the answer (e.g., ``\textit{how many days did Thurgood Marshall serve as a Supreme Court justice?}''). \\
\midrule
False Premise          & Questions that have a false preposition or assumption (e.g., ``\textit{What's the name of Taylor Swift's rap album before she transitioned to pop?}'' (Taylor Swift has not yet released any rap album)). \\ 
\bottomrule
\end{tabular}
\end{table}

CRAG contains two parts of data: the QA pairs and the contents for retrieval. We now describe each part of the data. Data generation details can be found in Appendix~\ref{appendix:kg_supported_qa}--\ref{appendix:mock}.

\subsection{Question answering pairs} 
CRAG covers five domains: Finance, Sports, Music, Movie, and Open domain, and eight types of questions, all in English. The question types are listed in Table~\ref{tab:question_type}. We constructed the question-answer pairs both from underlying KGs and web contents.

{\bf QA pairs constructed from KGs.}
\label{sec:kg_supported_qa}
We constructed QA pairs from KGs by collecting a set of entities based on publicly available data and then creating 600+ question templates based on selected entity types and relations. Next, we sampled entities with different popularities (head, torso and tail) following~\cite{sun2023head} from the KGs to fill in the templates and generate the full question and answer. %

{\bf QA pairs constructed from web contents.}
We asked annotators to write down possible questions that users may ask (e.g., ``\textit{most popular action movies in 2023}'') and created QA pairs from the corresponding web search results. %

Using the above methods, we collected 2,425 {\em Web Questions} and 1,984 {\em KG Questions}, with $661$, $658$, and $665$ {\em KG Questions} containing {\em head}, {\em torso}, and {\em tail} entities respectively. Tables~\ref{tab:data_statistics_by_dynamism}~and~\ref{tab:data_statistics_by_question_type} summarize the distribution of the questions across different dimensions. The size of each dimension slice (e.g., fast-changing facts) allows us to get metrics with $<5\%$ margin-of-error (with 95\% confidence level) for most of the cases. The dynamism distribution roughly reflects the nature of the domain (e.g., much more real-time questions for {\em Finance} than for other domains). See Appendix~\ref{appendix:definition_dynamism} for the definition of the dynamism categories.

\begin{table}[]
\centering
\small
\caption{The numbers and percentages (\%, in parenthesis) of questions for each category of dynamism, decided manually. Finance and Sports domain have the most Real-time and Fast-changing questions.}
\label{tab:data_statistics_by_dynamism}

\begin{tabular}{lrrrrrr}
\toprule
\textbf{Dynamism} & \textbf{Finance}  & \textbf{Sports}   & \textbf{Music}    & \textbf{Movie}    & \textbf{Open}     & \textbf{Total}    \\
\midrule
Real-time     & 434 (42)                   & 0 (\phantom{0}0)                    & 2 (\phantom{0}0)                    & 0 (\phantom{0}0)                    & 1 (\phantom{0}0)                    & 437 (10)                   \\
Fast-changing & 204 (20)                   & 275 (33)                   & 40 (\phantom{0}6)                    & 17 (\phantom{0}2)                    & 28 (\phantom{0}4)                    & 564 (13)                   \\
Slow-changing & 183 (18)                   & 215 (26)                   & 152 (24)                   & 253 (22)                   & 204 (26)                   & 1,007 (23)                   \\
Static        & 218 (21)                   & 343 (41)                   & 430 (69)                   & 855 (76)                   & 555 (70)                   & 2,401 (54)                   \\
\midrule
All         & 1,039 \phantom{(00)}            & 833 \phantom{(00)}              & 624 \phantom{(00)}                     & 1,125 \phantom{(00)}                 & 788 \phantom{(00)}                  & 4,409 \phantom{(00)}                    \\
\bottomrule
\end{tabular}
\end{table}

\begin{table}[]
\caption{The number and percentages (\%, in parenthesis) of questions for each question type, decided manually. Simple and simple with condition questions constitute $43\%$ of all questions.}
\label{tab:data_statistics_by_question_type}
\centering
\small

\begin{tabular}{lrrrrrr}
\toprule
\textbf{Question type} & \textbf{Finance}  & \textbf{Sports}   & \textbf{Music}    & \textbf{Movie}    & \textbf{Open}     & \textbf{Total}    \\
\midrule
Simple              & 466 (45)                   & 23 (\phantom{0}3)                    & 112 (18)                   & 519 (46)                   & 85 (11)                   & 1,205 (27)             \\
Simple w. condition & 113 (11)                   & 250 (30)                   & 92 (15)                   & 112 (10)                   & 122 (15)                   & 689 (16)             \\
Set                 & 48 (\phantom{0}5)                    & 93 (11)                   & 72 (12)                   & 104 (\phantom{0}9)                    & 86 (11)                   & 403 (\phantom{0}9)              \\
Comparison          & 146 (14)                   & 85 (10)                   & 102 (16)                   & 105 (\phantom{0}9)                    & 98 (12)                   & 536 (12)             \\
Aggregation         & 69 (\phantom{0}7)                    & 137 (16)                   & 96 (15)                   & 71 (\phantom{0}6)                    & 116 (15)                   & 489 (11)             \\
Multi-hop           & 86 (\phantom{0}8)                    & 64 (\phantom{0}8)                    & 55 (\phantom{0}9)                    & 90 (\phantom{0}8)                    & 87 (11)                   & 382 (\phantom{0}9)              \\
Post-processing heavy    & 26 (\phantom{0}3)                    & 24 (\phantom{0}3)                    & 26 (\phantom{0}4)                    & 28 (\phantom{0}2)                    & 76 (10)                   & 180 (\phantom{0}4)              \\
False Premise       & 85 (\phantom{0}8)                    & 157 (19)                   & 69 (11)                   & 96 (\phantom{0}9)                    & 118 (15)                   & 525 (12)             \\
\midrule
All                 & 1,039 \phantom{(00)}                    & 833 \phantom{(00)}              
     & 624 \phantom{(00)}                  & 1,125 \phantom{(00)}                & 788 \phantom{(00)}        
         & 4,409 \phantom{(00)} 
               \\
\midrule
\end{tabular}

\end{table}

\subsection{Contents for retrieval}
We included two types of contents for retrieval to simulate the practical scenario for RAG: web search and KG search.

{\bf Web search results.}
\label{sec:web_search_result}
For each question, we used the question text as the search query and stored up to $50$ HTML pages from the Brave search API~\cite{brave24}. See Table~\ref{tab:web_content_example} in Appendix~\ref{appendix:web_content_example} for an example. 

\revision{We estimated the web search recall with a heuristic-based method: first check whether the ground truth answer URL was found among the pages; if not, decide whether the fact in the ground truths is contained in the page snippet or content with an LLM. In particular, we pass the question, ground truth, and the pages to Llama 3 70B Instruct and ask it to judge if the context is sufficient to answer the question. }

\begin{figure}[t]
  \centering
  \includegraphics[width=0.8\linewidth]{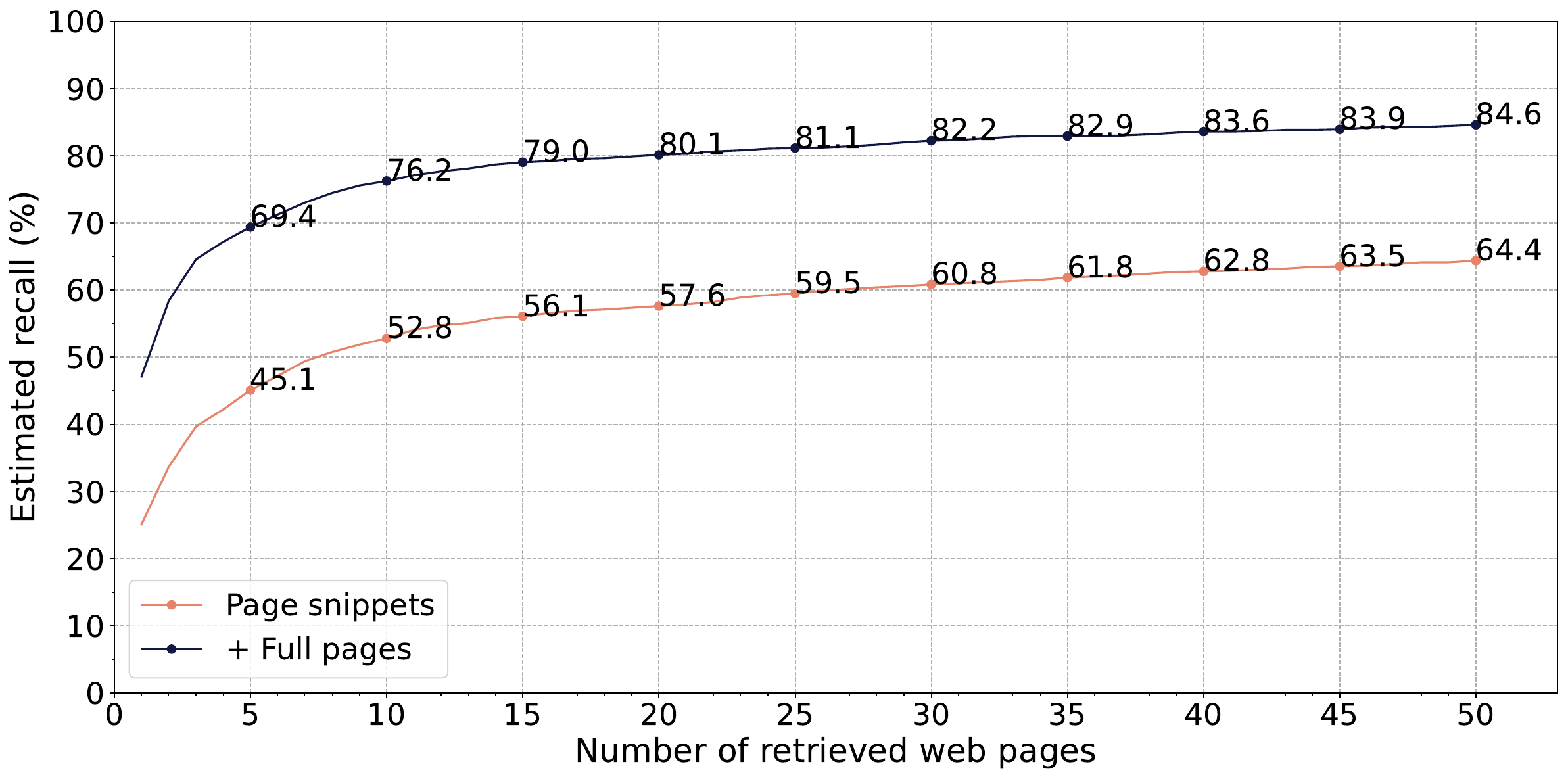}
  \caption{For $85\%$ of CRAG questions, the web search results are estimated to contain the ground truth facts. The curve shows that the retrieval recall grows sharply at the beginning and flattens out later on. %
}
  \label{fig:hits_at_k}
\end{figure}

\revision{Figure~\ref{fig:hits_at_k} shows the estimated web search recall curve for all CRAG questions, with an overall recall of 85\% when using all 50 pages.
It reflects multiple advantages of the benchmark by design.
First, the recall curves are sharp at the beginning and flatten out later, and the recall for the top 5 pages is about 69\%. This is comparable to what we observe in practice when developing a RAG system. The non-perfect coverage, especially for Task 1, allows us to test whether the RAG solutions admit ``I don't know'' when the retrieval results do not contain the necessary information. 
Second, compared to the recall from web snippets, full web pages increase the recall by about 20\%, emphasizing the importance of extracting and understanding HTML contents. Moreover, the estimated web search recall (50 web pages) is $93\%$ for {\em Web Questions} and $74\%$ for {\em KG Questions}, indicating significantly lower recall for KG questions than web questions. 
This aligns with our observations that web search recall for torso and tail entities is typically lower, underlining the crucial role of leveraging KGs in Task 2 and 3.} %

{\bf Mock KGs.}
We created mock KGs that contain publicly available KG data used to generate the questions, randomly selected entities of the same type, and also ``hard negative'' entities with similar names (e.g., \textit{``phantom''} for \textit{``phantom of the opera''}).

{\bf Mock APIs.}
We created mock APIs with pre-defined parameters to support structured search in the mock KGs. For example, for queries asking for stock prices, an example mock API is in the form of \verb|get_price_history(ticker)|. 

\smallskip
\revision{We collected snapshots of the KG and web search data concurrently while posing real-time and fast-changing questions. This approach ensures that we capture the ``snapshot'' of the information world at the time of question answering. A RAG solution that performs well on the benchmark should also be capable of reasoning over time and generalizing to evolving questions.}

In total, the resulting data contains 220K webpages, a KG of 2.6M entities, and 38 mock APIs. \revision{See Table~\ref{tab:mockapis} in the Appendix for a complete list of the mock APIs.} %

\section{Metrics and Evaluation}
\label{sec:kdd_cup}
In this section, we present the metrics for evaluating the RAG systems and briefly describe the 2024 Meta KDD Cup challenge in Appendix~\ref{appendix:evaluation_test_sets}.

\subsection{Metrics}
\label{sec:metrics}
We use a scoring method to assess the performance of RAG systems. For each question in the evaluation set, we first label the answer with \textbf{perfect, acceptable, missing,} or \textbf{incorrect}, according to the following criteria.

\textbf{Perfect.} The response correctly answers the user's question and contains no hallucinated content.

\textbf{Acceptable.} The response provides a useful answer to the user's question but may contain minor errors that do not harm the usefulness of the answer.

\textbf{Missing.} The response is ``I don't know'', ``I'm sorry I can't find ...'', a system error such as an empty response, or a request from the system to clarify the original question.

\textbf{Incorrect.} The response provides wrong or irrelevant information to answer the user's question.
 
We use a scoring method with score $1$, $0.5$, $0$, and $-1$ for each \textit{perfect, acceptable, missing}, and \textit{incorrect} answer, respectively, where we penalize hallucinated answers and prefer \textit{missing} answers to \textit{incorrect} ones. \revision{We then define \textbf{truthfulness} as the average score from all examples in the evaluation set for a given RAG system.}

\subsection{Evaluation}
\label{sec:evaluation}
Similar to previous work~\cite{lfqa23}, we employ both human evaluation {\bf (human-eval)} and model-based automatic evaluation {\bf (auto-eval)}. In the former, we use manual grading to judge {\em perfect, acceptable, missing}, and {\em incorrect} for each answer. In the latter, we merge {\em perfect} and {\em acceptable}, call it {\bf accurate}, and use a three-way scoring system with $1,-1,0$ for {\em accurate}, {\em incorrect}, and {\em missing} answers. 

We design a two-step method for automatic evaluation: if the answer matches the ground truth exactly, it is considered {\em accurate}; otherwise, we use LLMs to determine whether the response is {\em accurate, incorrect}, or {\em missing}. To avoid the {\em self-preference} problem~\cite{panickssery2024llm}, we use two LLM evaluators: ChatGPT (\texttt{gpt-3.5-turbo-0125})~\cite{chatgpt2023} and Llama 3 (\texttt{llama-3-70B-instruct})~\cite{llama3modelcard} and report the average {\em accurate, hallucination, missing} rates, and \revision{{\em truthfulness}} scores from the two models for each RAG system. Our offline experiment shows that this two-step method yields an average F1 score of $94.7\%$ for ChatGPT and $98.9\%$ for Llama 3 compared to human-eval. See Appendix~\ref{appendix:auto-eval} for more details.

{\bf Test data split.}
We split the data randomly into {\em validation} (30\%), {\em public test} (30\%), and {\em private} (40\%), and released the validation and public test sets (Appendix~\ref{appendix:evaluation_test_sets}). Participants of the KDD Cup challenge can use the validation and public test sets to develop and test their models, and the submitted solutions were evaluated on the private test set. Future offline users of CRAG can use the validation set for development, fine-tuning, and validation, and the public test set for testing and result reporting.

\section{Benchmarking}
\label{sec:benchmarking}
In this section, we present the performance of LLMs and RAG systems on CRAG, demonstrating that CRAG has a reasonable level of difficulty and can help draw insights and show directions in developing RAG techniques.

\subsection{Straightforward RAG solutions}
\label{subsec:baselines}

{\bf Experiment setup:} We started with running LLM-only solutions on the CRAG public test set with $1,335$ questions, using simple prompts that encourage brief answers and {\em ``I don't know''} answers when the confidence is low (Appendix~\ref{appendix:prompt}). We employed Llama 2 Chat (\texttt{llama-2-7b-chat} and \texttt{llama-2-70b-chat})~\cite{touvron2023llama}, Llama 3 Instruct (\texttt{llama-3-8B-instruct} and \texttt{llama-3-70B-instruct})~\cite{llama3modelcard}, \revision{Mixtral (\texttt{Mixtral-8x7B-Instruct-v0.1})~\cite{jiang2023mistral}, Falcon (\texttt{40B})~\cite{almazrouei2023falcon},  FLAN-T5 (\texttt{FLAN-T5-XXL})~\cite{chung2024scaling}, and GPT-4 Turbo (\texttt{gpt-4-turbo-2024-04-09})~\cite{achiam2023gpt}}. The web-only RAG solutions we evaluated (Task 1) used a fixed-length web context window (\revision{1K tokens for Falcon and FLAN-T5}, 2K for Llama 2 Chat, and 4K for Llama 3 Instruct and GPT-4 Turbo); we concatenated webpage snippets using the original order from the data as the reference text, until filling up the window (similar to~\cite{vu2023freshllms, kandpal2023large, mallen2023}). Our KG-based solutions (Tasks 2, 3) additionally used a fixed-length KG context window (\revision{0.5K tokens for Falcon and FLAN-T5}, 1K for Llama 2 Chat, and 2K for Llama 3 Instruct and GPT-4 Turbo) to include the results from the mock APIs; we extracted the relevant query entities using \texttt{llama-3-8B-instruct} with in-context learning (similar to~\cite{Schick2023ToolFormer}) detailed in Appendix~\ref{appendix:prompt} and concatenated the results returned from all applicable mock APIs (based on the extracted entities), until filling up the window. We provide an extensive comparison of all LLMs in Appendix~\ref{appedix:baseline_benchmarking} and focus on the best-performing LLMs (i.e., Llama 3 70B Instruct and GPT-4 Turbo) in this section.

\begin{table*}[t]
  \caption{Performance of straightforward RAG solutions. All numbers are in percentage. LLM-only solutions has up to 34\% accuracy and straightforward RAG solutions has up to 44\% accuracy. \revision{The subscript in ``$\text{truthfulness}_a$'' denotes the result is reported by auto-eval.}}
  \label{tab:baseline_benchmarking}
\centering
\small
\begin{tabular}{llrrrrr}
\toprule
                  & \textbf{Model} & \textbf{Accuracy} & \textbf{Hallucination} & \textbf{Missing} & \revision{\textbf{Truthfulness$_a$}} \\
\midrule
\textbf{LLM only} & Llama 3 70B Instruct   & 32.3                   & 28.9                        & 38.8                  & 3.4                 \\
                  & GPT-4 Turbo          & \textbf{33.5}          & \textbf{13.5}               & 53.0                  & \textbf{20.0}       \\
\midrule
\textbf{Task 1}                     & Llama 3 70B Instruct   & 35.6                   & 31.1                        & 33.3                  & 4.5                 \\
                  & GPT-4 Turbo          & \textbf{35.9}          & \textbf{28.2}               & 35.9                  & \textbf{7.7}            \\
\midrule
\textbf{Task 2}                     & Llama 3 70B Instruct   & 37.5                   & 29.2                        & 33.3                  & 8.3                  \\
                  & GPT-4 Turbo          & \textbf{41.3}          & \textbf{25.1}               & 33.6                  & \textbf{16.2}            \\
\midrule
\textbf{Task 3}   & Llama 3 70B Instruct   & 40.6                   & 31.6                        & 27.8                  & 9.1                                         \\
                  & GPT-4 Turbo          & \textbf{43.6}          & \textbf{30.1}               & 26.3                  & \textbf{13.4}              \\
\bottomrule
\end{tabular}
\end{table*}

\begin{figure}[t]
  \centering
  \includegraphics[width=\linewidth]{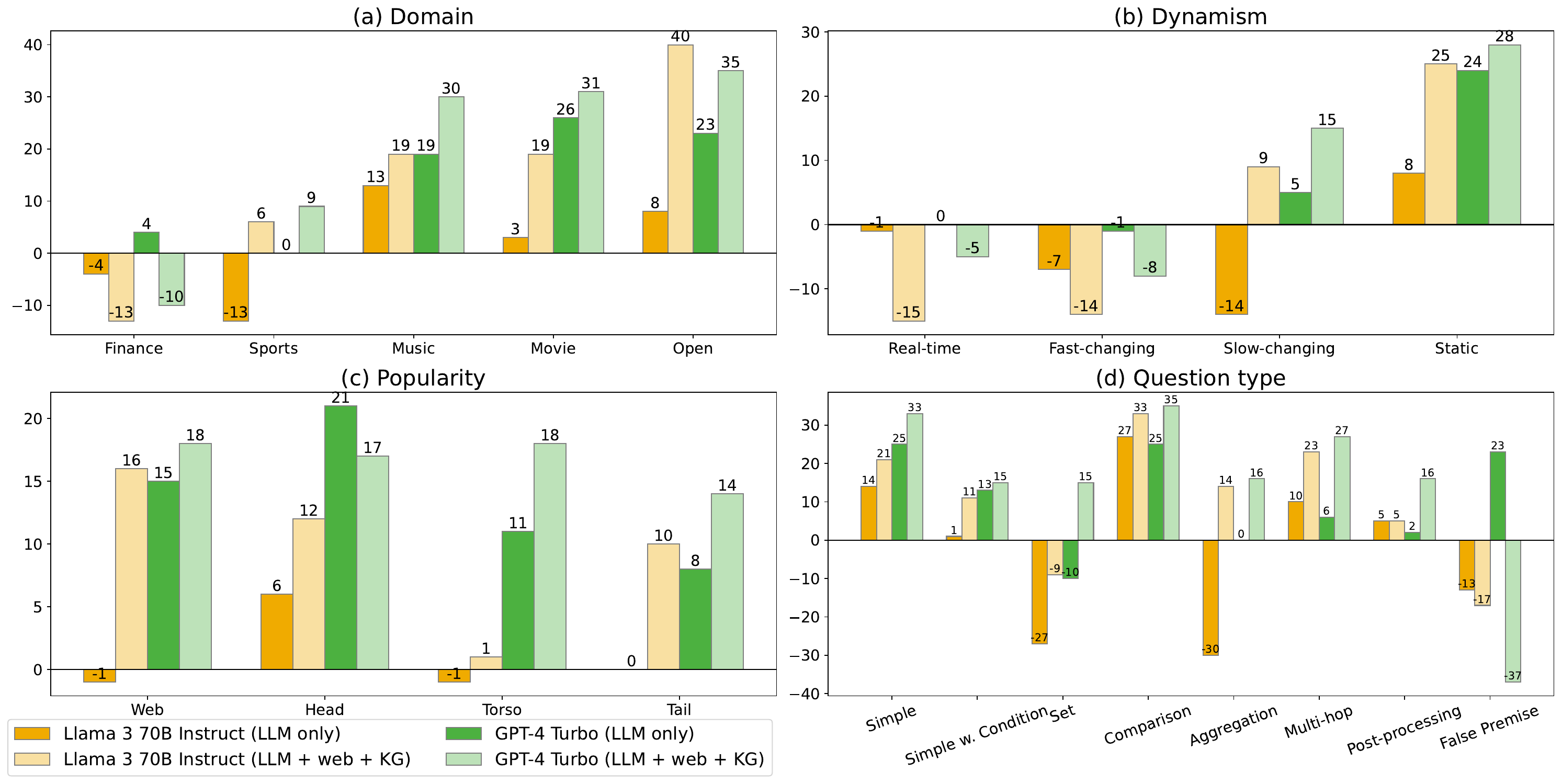}
  \caption{LLM-only and Task 3 solution auto-eval truthfulness (in percentage) across domain, dynamism, popularity, and question type.}
  \label{fig:baseline_slicing}
\end{figure}

Table~\ref{tab:baseline_benchmarking} shows the average evaluation results from the two auto-evaluators (ChatGPT and Llama 3) and illustrates that the CRAG benchmark is {\bf\em non-trivial}. First, the best LLM-only solutions (GPT-4 Turbo) obtained an accuracy of only 34\%, with \revision{truthfulness} of 20\%, showing a big room for improvement. Second, straightforward RAG solutions obtained up to 44\% accuracy, showing that extra information {\em does} help answer more questions reliably. Interestingly, none of the RAG solutions obtain \revision{truthfulness} higher than 20\%; this is because all RAG solutions introduce more hallucinations generated from irrelevant retrieval results, showing a big challenge in RAG---{\em How to judiciously use retrieval results without being distracted by retrieval noises?} Third, we found that Task 2 \revision{truthfulness scores} are higher than Task 1, showing that the KG knowledge helps improve accuracy, with a similar or even lower hallucination rate, because the KG knowledge is typically brief but precise. Unfortunately, the improvement is mediocre, showing a second challenge in RAG---{\em How to best leverage the power of KG data?} Finally, the \revision{truthfulness} for Task 3 are also higher than Task 2, because of better search ranking (recall that Task 1 and 2 provide five pages randomly selected from the top-10 search results) and better search recall. \revision{In particular, we found that the ground truths of over 30\% of questions are available in the web retrieval results but are not included in the prompt due to the context window limitation.} This shows {\em the importance of search ranking} in RAG.

Figure~\ref{fig:baseline_slicing} shows the auto-eval results across the domain, dynamism, popularity, and question type dimension. The results reveal a lot of interesting observations and show that the CRAG benchmark allows more {\bf\em insightful} conclusions. First, it shows {\em which slices of the benchmark are harder}. For example, we found much lower RAG \revision{truthfulness} on the {\em Finance} and {\em Sports} domains, for {\em real-time} and {\em fast-changing} facts, for {\em tail} entities, and for complex questions requiring {\em set answers, post-processing,} and with {\em false premises}. Second, it shows {\em where it is harder to leverage retrieval results}. Take the popularity slices as an example, we observed that GPT-4 Turbo's \revision{truthfulness} dropped from head (21\%) to torso (11\%) to tail (8\%), consistent with past observations~\cite{sun2023head}; however, the straightforward RAG solution based on GPT-4 Turbo improved QA quality regarding torso (+7\%) %
and tail entities (+6\%) but lowered the quality regarding head (-4\%). 
Finally, although our goal is {\em not} to compare different LLMs, the different dimensions allow us to understand the strengths and weaknesses of each method. For example, although the RAG system based on Llama 3 70B Instruct has a lower overall \revision{truthfulness score} than the one based on GPT-4 Turbo, it has a similar or slightly higher \revision{truthfulness} in answering {\em simple} and {\em comparison} questions, whereas much lower \revision{truthfulness} in answering {\em set} and {\em post-processing} questions, suggesting investigations on the reasoning capabilities.

\subsection{State-of-the-art industry solutions}
Next, we evaluated industry state-of-the-art (SOTA) RAG solutions on CRAG public test set. We selected five RAG systems built upon SOTA LLMs and search engines, queried them with CRAG questions, collected the responses, and applied manual grading (details in Appendix~\ref{appedix:sota_benchmarking}). %

In addition, we applied traffic weights to the questions to understand the solutions in real-world use cases. \revision{The traffic weights reflect the frequency of each question type, as defined in Table~\ref{tab:question_type}, in real QA traffic. We gave the same weights to all domains and reported the macro average across domains. This is because we have observed quite different domain-level distributions in different use cases, but have been observing similar distributions at the query-type level.}

\begin{table*}[t]
  \caption{Benchmarking CRAG questions with industry SOTA RAG systems. Perfect, acceptable (Acc.), hallucination (Hall.), missing rates (Miss.), and \revision{truthfulness$_h$ reported by human-eval (Truth$_h$)} are in percentages. The best system achieves \revision{truthfulness} of 51\% and provides perfect answers for up to 63\% of questions.}
  \label{tab:sota_benchmarking}
  \centering
  \small
\begin{tabular}{llrrrrrr}
\toprule
                                          & \textbf{System}                            & \textbf{Perfect}     & \textbf{Acc.}  & \textbf{Hall.} & \textbf{Miss.}     & \textbf{\revision{Truth}$_h$}       & \textbf{Latency (ms)} \\
\midrule
\textbf{Equal}    & Copilot Pro                                & \textbf{62.6}    & 11.7                & 17.9                   & 7.8              & \textbf{50.6} & 11,596                 \\
\textbf{weighted} & Gemini Advanced                            & 60.8             & 10.1                & 16.6          & 12.5             & 49.3          & 5,246                  \\
                  & ChatGPT Plus                               & 59.8             & \textbf{13.3}       & 25.0                     & 1.9              & 41.5          & 6,195                  \\
                  & Meta SG                                    & 52.5             & 9.7                 & \textbf{16.0}                     & 21.8             & 41.4           & \textbf{3,431}   \\
                  & Perplexity.ai & 55.8             & 8.8                 & 25.3                   & 10.1             & 34.9           & 4,634         \\
\midrule
\textbf{Traffic}  & Copilot Pro  & \textbf{70.0} & 9.5           & 14.3                             & 6.1  & \textbf{60.5}  & - \\
\textbf{weighted} & Gemini Advanced           & 67.1          & 10.0          & \textbf{12.7} & 10.2 & 59.3          & - \\
                  & ChatGPT Plus  & 61.8          & \textbf{11.4} & 25.7          & 1.3  & 41.8          & - \\
                  & Meta SG                   & 61.0          & 7.1           & 14.1          & 17.8 & 50.5          & - \\
                  & Perplexity.ai             & 63.7          & 6.3           & 20.9          & 9.1  & 45.9          & - \\

\bottomrule
\end{tabular}
\end{table*}

\begin{figure}[t]
  \centering
  \includegraphics[width=\linewidth]{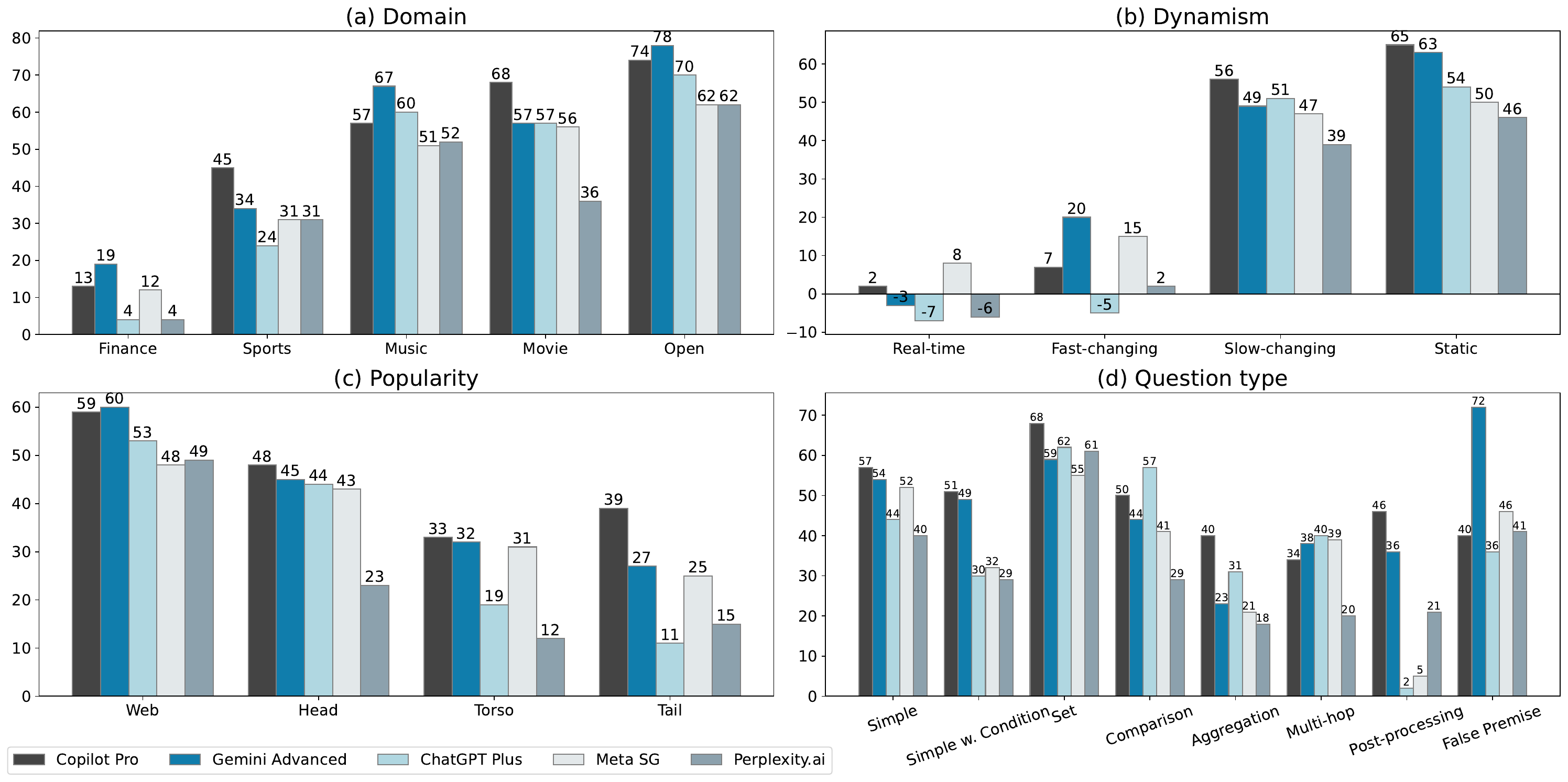}%
  \caption{SOTA systems human-eval \revision{truthfulness scores} (in percentage) across different dimensions. 
  }
  \label{fig:sota_slicing}
\end{figure}

Table~\ref{tab:sota_benchmarking} and Figure~\ref{fig:sota_slicing} show the overall performance of the SOTA systems and their performance across different dimensions. The evaluation results confirm our belief that the CRAG benchmark {\em reveals interesting insights and shows room for improvement for existing RAG solutions.} 
First, the results from SOTA solutions achieve much better \revision{truthfulness} (highest $51\%$) compared to the straightforward solutions. However, the hallucination rate ranges from \revision{16\%} to 25\%, so the answers are still {\em not} trustworthy. Note that the \revision{truthfulness scores} between the SOTA solutions and the straightforward solutions are not completely comparable, as they have different accesses to retrieval contents (Appendix~\ref{appendix:baseline_benchmarking} and \ref{appedix:sota_benchmarking_quality}), and the former used auto-eval, while the latter used human-eval; however, the trend is valid. 
\revision{Second, we observed very different latency, ranging from $3.4$s to $11.6$s, reflecting the different design options in trading off latency and quality; for example, Copilot Pro has the highest truthfulness, but meanwhile highest latency, whereas Meta SG~\cite{xin2024journey} has mid-tier truthfulness but lowest latency. (See Appendix~\ref{appedix:sota_benchmarking_latency} for additional results and how we measured latency.)
Third,} most difficult slices we see in the straightforward solutions remain to be difficult for SOTA solutions: {\em real-time} and {\em fast-changing} queries, and questions regarding {\em torso} and {\em tail} entities, showing the improvement needed for handling retrieval noises when the system relies on retrieval results to answer the question; as another example, we see lower \revision{truthfulness} for queries requiring {\em aggregation, multi-hop reasoning} or {\em post-processing}, showing the improvement space for reasoning in question answering. 
Last, \revision{truthfulness} on {\em set} and {\em false premise} questions improved significantly in the SOTA solutions compared to the straightforward solutions, showing advancement in RAG systems in providing accurate and complete set answers and detecting {\em false premises}.

\section{Conclusion}
\label{sec:conclusion}

This paper proposes \sys, a rich and comprehensive benchmark designed to advance research in retrieval-augmented generation (RAG). With detailed empirical studies, \sys reviewed gaps in existing RAG solutions and provided valuable insights for future improvement. 
We plan to continue improving and expanding the benchmark for multi-lingual questions, multi-modal questions, multi-turn conversations, etc., to ensure \sys stays at the forefront to push RAG research, adapts to emerging challenges, and evolves for new research needs.

\section*{Acknowledgements}
We would like to thank Jinsong Yu for his advice on the data strategy and help in connecting with the partner teams.
We thank Alex Boesenberg for enabling us to create web search results. We thank Sam Wexler for coordinating this project with various partners. We thank Hejia Zhang, Rares Bostan, Eryk Helenowski, Nanshu Wang, Sinong Wang, and Denis Savenkov for the discussion regarding LLM and RAG evaluation. We thank Katie Manlove for coordinating with the budget and annotation needs. We thank our annotation team for creating many great examples: David Vu, Florian Gawlitta, Rani Salomi Pitta, Lauren Gregory Mikus, Tara Welch, Nidhi Modi, Ray De Leon, Jader Ricarte, Joshua Aranzaso. We thank Apoorva Srinivas and Courtnee Parker for coordinating the annotation tasks. We would like to acknowledge the support from Jackie Leader, Mindy Abern, Heather Nolan, Ty Toledano, George Lan, Ben Edgar, Sy Choudhury, Rodrick Shepard, Katie Manlove, Mavis Hu, Jen Vinz, Taha Masood, Parkin Kent, Rebekkah Hogan, Tony Nelli, Jennifer Pak, Jonathan Torres, and Amy Lee. 

Lei Chen's work is partially supported by National Key Research and Development Program of China Grant No. 2023YFF0725100, National Science Foundation of China (NSFC) under Grant No. U22B2060, the Hong Kong RGC GRF Project 16213620, RIF Project R6020-19, AOE Project AoE/E-603/18, Theme-based project TRS T41-603/20R, CRF Project C2004-21G, Guangdong Province Science and Technology Plan Project 2023A0505030011, Hong Kong ITC ITF grants MHX/078/21 and PRP/004/22FX, Zhujiang scholar program 2021JC02X170, Microsoft Research Asia Collaborative Research Grant, HKUST-Webank joint research lab and HKUST(GZ)-Chuanglin Graph Data Joint Lab.

\bibliographystyle{abbrv}
\bibliography{reference}

\newpage
\appendix
\clearpage
\section{Appendix}

\smallskip
\noindent

\subsection{Dataset}
\label{appendix:dataset}
\subsubsection{Constructing QA pairs from KGs}
\label{appendix:kg_supported_qa}
We first collected a set of entities based on publicly available data. Then we created question-answer pairs in three steps for {\em Simple static and dynamic questions}. 

\noindent
{\em Step 1.} For each domain, we first selected an entity type and a meaningful relation $(e, r)$ and created a question template. For example, for \textit{(music artist, first album)}, we create a template ``\textit{what is the first album of [music artist]?}''. 

\noindent
{\em Step 2.} We then sampled entities from the KGs to fill in the templates and generate the full question. We adopted the method described in~\cite{sun2023head} and sampled entities of top, middle, and bottom popularity. We defined popularity based on heuristics for each entity type and created an equal number of questions for each bucket.

\noindent
{\em Step 3.} 
Last, we took the associated attribute values as the answer to the question to create question-answer pairs. 

\smallskip
We created the {\em Comparison, Aggregation, Set, Post-processing}, and {\em False-premise questions} in a similar way but 1) made sure the template allows for crisp and deterministic answers and 2) sampled the subject entities that fit the question. We used heuristics to select entity types for these question categories.

Finally, we created multi-hop questions in three steps, similar to those described in~\cite{talmor2018web}. We first sampled an entity $e_1$ from the KG and selected two relation triplets following a two-hop path: $(e_1, r_1, e_2)$ and $(e_1, r_2, e_3)$. We then created a question template describing the path. %
For example, for path \textit{(company$_1$, is\_parent, company$_2$)} followed by \textit{(company$_1$, ceo, person)}, we created the template \textit{"who is the CEO of the parent company of [company$_2$]?"}. The answer to the new question will be $e_3$ in the second triplet.

\subsubsection{Definition of dynamism categories}
\label{appendix:definition_dynamism}

\begin{table}[h!]
\centering
\small
\caption{Definition of dynamism categories.}
\begin{tabular}{ll}
\toprule
\textbf{Dynamism} & \textbf{Definition}                                                                                                                        \\
\midrule
Real-time         & \begin{tabular}[c]{@{}l@{}}The answer to the question changes over seconds \\ (e.g., ``\textit{What's Costco's stock price today?}'').\end{tabular}      \\
\midrule
Fast-changing     & \begin{tabular}[c]{@{}l@{}}The answer to the question changes no more than daily \\ (e.g., ``\textit{When is Laker's game tonight?}'').\end{tabular}         \\
\midrule
Slow-changing     & \begin{tabular}[c]{@{}l@{}}The answer to the question changes no more than yearly \\ (e.g., ``\textit{Who won the Grammy award last year?}'').\end{tabular} \\
\midrule
Static            & \begin{tabular}[c]{@{}l@{}}The answer to the question does not change over time, \\ such as the birth date of a person.\end{tabular}            \\
\bottomrule
\end{tabular}
\end{table}

\subsubsection{Constructing QA pairs from web contents}
\label{appendix:web_supported_qa}
\noindent
{\em Step 1.} Ask annotators to write down a list of questions that could possibly be answered by web search based on a general guideline (e.g., ``\textit{what is the most popular action movie in 2023?}''). %

\noindent
{\em Step 2.} Generate the web search results to answer the question. %

\noindent
{\em Step 3.} Finally, annotators reviewed the web search results to determine the ground truth answers to the questions:
  1) If the search results successfully provided the necessary information, annotators recorded the ground truth answer text and the URL associated with it based on the retrieved content. Note that the answer is determined by the \textit{query\_time} at which the web search happened, especially for the \textit{Fast-changing} and \textit{Real-time} questions.
  2) Otherwise, annotators conducted further web searches to document the correct answers.

Besides the QA pairs, the annotators will also provide labels for the domain, dynamism, question types, and an {\em answer URL} (a URL that contains the answer to the question) for {\em Web Questions}.

\subsubsection{Validation for the QA pairs}
We conducted two phases of dataset validation with our in-house Linguist team.

\noindent \textbf{Phase 1. Question and meta-label validation.} After the initial round of QA pair generation, an audit session was conducted, where expert annotators reviewed the question template, questions, and meta-labels (domain, question type, etc), applying edits as necessary with 2x human review (agreement rate $90\%+$). All problematic questions (e.g., a wrong false-premise question) were revised, and all conflicting labels were resolved by a third more experienced auditor.

\noindent \textbf{Phase 2. Answer validation.} To ensure the answers in the benchmark are correct, we further conducted an auditing process for all the answers.
For the web questions, an annotation team reviewed each question and conducted an extensive search to make sure the answer is factually correct and includes comprehensive information (such as for set questions) with 2x human review (agreement rate $90\%+$). A third more experienced auditor then reviewed all conflicting answers and provided a resolution. 
For the KG questions, a team of five engineers carefully checked the questions and queried the mock APIs manually to validate the answers. This step resulted in a $5\%$ answer correction.

\noindent In both phases, we paid special attention to examples where the straightforward solutions output different answers from the ground truth answers and asked the auditing team to double-check those examples. This step yielded an additional $2\%$ answer updates.

\subsubsection{An example of retrieved web search results}
\label{appendix:web_content_example}

\begin{table*}[h!]
  \caption{An example of web search results.}
  \label{tab:web_content_example}
  \centering
  \small
  \begin{tabular}{p{2.5cm}p{10.6cm}}
    \toprule
    Key & Value \\
    \midrule
    "page name" & "A Short History Of ChatGPT: How We Got To Where We Are Today"  \\
    "page url" & "https://www.forbes.com/sites/bernardmarr/2023/05/19/a-short-history-of-chatgpt-how..." \\
    "page snippet" & "OpenAI released an early demo of ChatGPT on <strong>November 30, 2022</strong>..." \\
    "page last modified" & "2024-1-18 15:32:24" \\
    "html page" & "<!DOCTYPE html><html lang="en"><head><link rel="preload" as="font" href="https..." \\
    \bottomrule
  \end{tabular}
\end{table*}

\subsubsection{The mock data and mock APIs}
\label{appendix:mock}
CRAG provides mock APIs to simulate retrieval from web, KG, and real-time APIs in the {\bf\em real} retrieval environment, allowing {\bf\em accessible} facilitating data and fair comparison. 

CRAG provides both structured (through mock KGs) and unstructured (web search results) information to test the effectiveness of RAG systems in leveraging a diverse range of available information. First, for each question in the benchmark, CRAG provides up to 50 web search results from a real-world search engine---the Brave Search API~\cite{brave24}. Different from existing benchmarks that use snippets or selected text chunks~\cite{bajaj2018ms,chen2023benchmarking}, CRAG provides full HTML pages, containing more information and potentially more noises as in a realistic setting. Second, CRAG provides mock KG search APIs to test structured search for RAG. The mock KGs, though much smaller in size, contain both information necessary to answer a subset of questions in the benchmark and noises that have similar entity or attribute names, again simulating real settings. Our mock KGs contain about 2.6M entities and have a signal-to-noise ratio of less than 1/30.

\begin{table}[h!]
\centering
\scriptsize
\caption{Mock APIs and Descriptions.}
\label{tab:mockapis}
\begin{tabular}
{p{0.55\textwidth}p{0.38\textwidth}}
\toprule
\textbf{APIs} & \textbf{Descriptions}\\
\midrule
open\_search\_entity\_by\_name(query: str) -> dict & Search for entities by name in the Open domain.\\
open\_get\_entity(entity: str) -> dict & Retrieve detailed information about an entity in the Open domain.\\
\midrule
movie\_get\_person\_info(person\_name: str) -> dict & Get information about a person related to movies.\\
movie\_get\_movie\_info(movie\_name: str) -> dict & Get information about a movie.\\
        movie\_get\_year\_info(year: str) -> dict & Get information about movies released in a specific year.\\
        movie\_get\_movie\_info\_by\_id(movie\_id: int) -> dict & Get movie information by its unique ID.\\
        movie\_get\_person\_info\_by\_id(person\_id: int) -> dict & Get person information by their unique ID.\\
        \midrule
        finance\_get\_company\_name(query: str) -> dict & Search for company names in the finance domain.\\
        finance\_get\_ticker\_by\_name(query: str) -> dict & Retrieve the ticker symbol for a given company name.\\
        finance\_get\_price\_history(ticker\_name: str) -> dict & Get the price history for a given ticker symbol.\\
        finance\_get\_detailed\_price\_history(ticker\_name: str) -> dict & Get detailed price history for a ticker symbol.\\
        finance\_get\_dividends\_history(ticker\_name: str) -> dict & Get dividend history for a ticker symbol.\\
        finance\_get\_market\_capitalization(ticker\_name: str) -> dict & Retrieve market capitalization for a ticker symbol.\\
        finance\_get\_eps(ticker\_name: str) -> dict & Get earnings per share (EPS) for a ticker symbol.\\
        finance\_get\_pe\_ratio(ticker\_name: str) -> dict & Get the price-to-earnings (PE) ratio for a ticker symbol.\\
        finance\_get\_info(ticker\_name: str) -> dict & Get financial information for a ticker symbol.\\
        \midrule
        music\_search\_artist\_entity\_by\_name(artist\_name: str) -> dict & Search for music artists by name.\\
        music\_search\_song\_entity\_by\_name(song\_name: str) -> dict & Search for songs by name.\\
        music\_get\_billboard\_rank\_date(rank: int, date: str = None) -> dict & Get Billboard ranking for a specific rank and date.\\
        music\_get\_billboard\_attributes(date: str, attribute: str, song\_name: str) -> dict & Get attributes of a song from Billboard rankings.\\
        music\_grammy\_get\_best\_artist\_by\_year(year: int) -> dict & Get the Grammy Best New Artist for a specific year.\\
        music\_grammy\_get\_award\_count\_by\_artist(artist\_name: str) -> dict & Get the total Grammy awards won by an artist.\\
        music\_grammy\_get\_award\_count\_by\_song(song\_name: str) -> dict & Get the total Grammy awards won by a song.\\
        music\_grammy\_get\_best\_song\_by\_year(year: int) -> dict & Get the Grammy Song of the Year for a specific year.\\
        music\_grammy\_get\_award\_date\_by\_artist(artist\_name: str) -> dict & Get the years an artist won a Grammy award.\\
        music\_grammy\_get\_best\_album\_by\_year(year: int) -> dict & Get the Grammy Album of the Year for a specific year.\\
        music\_grammy\_get\_all\_awarded\_artists() -> dict & Get all artists awarded the Grammy Best New Artist.\\
        music\_get\_artist\_birth\_place(artist\_name: str) -> dict & Get the birthplace of an artist.\\
        music\_get\_artist\_birth\_date(artist\_name: str) -> dict & Get the birth date of an artist.\\
        music\_get\_members(band\_name: str) -> dict & Get the member list of a band.\\
        music\_get\_lifespan(artist\_name: str) -> dict & Get the lifespan of an artist.\\
        music\_get\_song\_author(song\_name: str) -> dict & Get the author of a song.\\
        music\_get\_song\_release\_country(song\_name: str) -> dict & Get the release country of a song.\\
        music\_get\_song\_release\_date(song\_name: str) -> dict & Get the release date of a song.\\
        music\_get\_artist\_all\_works(artist\_name: str) -> dict & Get all works by an artist.\\
        \midrule
        sports\_soccer\_get\_games\_on\_date(team\_name: str, date: str) -> dict & Get soccer games on a specific date.\\
        sports\_nba\_get\_games\_on\_date(team\_name: str, date: str) -> dict & Get NBA games on a specific date.\\
        sports\_nba\_get\_play\_by\_play\_data\_by\_game\_ids(game\_ids: List[str]) -> dict & Get NBA play by play data for a set of game ids.\\

\bottomrule
\end{tabular}
\end{table}

\subsection{Evaluation}
\label{appendix:evaluation}
\subsubsection{Human evaluation}
\label{appendix:human-eval}
We run human evaluation to score each answer with respect to the metrics defined in Section \ref{sec:metrics}. We score each \textit{perfect, acceptable, missing}, or \textit{incorrect} answer with a score $s_p$, $s_a$, $s_m$, and $s_{in}$, respectively and define \revision{{\bf truthfulness$_h$}} as the score of the answer by setting $s_p = 1$, $s_a = 0.5$, $s_m = 0$, and $s_{in} = -1$. We then compute the average \revision{truthfulness} for all examples in the evaluation set as the \revision{truthfulness} score for the RAG solution. %

\revision{{\bf Human-eval instructions.} The human-eval instructions are as follows.\\
Given a query, a query day and time at which the query was made, the chatbot’s response, grade the \textit{accuracy} for the response according to the criteria below:\\
Using an external search engine, please evaluate the factual accuracy of the response based on the grading rubric below. An accurate answer should be factually correct and provide useful information to answer the user’s question.
\begin{itemize}
    \item Accuracy = 0 (Missing). It covers the following situations. 
        There is no response. 
        There is a failure to provide a response to the request (e.g. “I’m sorry ……, I can’t do …”) due to inability. 
        E.g., 
            Query: “Latest news about the Nobel prize today.”
            Response: “I can’t find specific information regarding the Nobel prize …”
        The response fails to answer and asks follow-up questions.
    \item Accuracy = 1 (Incorrect). It covers the following situations. 
        The answer is unintelligible.
        The answer is poorly formed.
        The answer contains a major hallucination (e.g. wrong date, wrong numbers, or other significant factual errors).
        The answer is irrelevant to the user’s request.
        Answers in a category, location, or time window that is significantly different from the user’s request, if any.
        There is a significant structural/formatting error.
        The response is otherwise a total structural/functional failure and does not contain sufficient well-formed content that can be used to determine accuracy
    \item Accuracy = 2 (Acceptable). It covers the following situations. 
        The answer is acceptably correct and relevant to the user's request, but may miss some information, i.e. accurate but not complete, or mostly accurate with minor issues. 
        The answer may contain some minor hallucination that doesn’t significantly alter the overall meaning. “Minor hallucination” means the answer addressed the user's question but might be off on some additional details. The rule of thumb is to see if the answer serves the purpose of the user’s question, and whether the hallucination could mislead the users on what they were asking. 
    \item Accuracy = 3 (Accurate).
        The answer is factually correct, contains all the relevant information requested, responds appropriately to the query, and does not contain any hallucination.
\end{itemize}
}

\revision{We requested two human graders to grade each question (agreement rate $94\%$), and when there is a conflict, a third more experienced grader will resolve it.} 

\subsubsection{Automatic evaluation}
\label{appendix:auto-eval}
In auto-eval, we merge \textit{perfect} and \textit{acceptable} as \textit{accurate} and consider only three scores: 1 for {\em accurate}, 0 for {\em missing}, and -1 for {\em incorrect}. The \revision{{\bf truthfulness$_a$}} is calculated by the average \revision{truthfulness} for all the examples in the evaluation set, and is effectively  
$$ \text{accuracy} - \text{hallucination},$$
where \textbf{accuracy}, \textbf{hallucination}, and \textbf{missing} are the percentage of \textit{accurate}, \textit{incorrect}, and \textit{missing} answers in the test set. These score choices penalize \textit{incorrect} answers, award \textit{correct}, and assign a value of 0 to \textit{missing} answers. 

{\bf Auto-evaluators.} We computed the accuracy and F1 for the two auto-evaluators for the {\em accurate, incorrect}, and {\em missing} examples in the public test set, respectively. Here, we considered human-eval labels as the ground truth. Table~\ref{tab:auto-eval_accuracy} shows both models attain reasonable accuracy and F1 scores as an evaluator compared to human evaluation. 

\begin{table}[]
\caption{Accuracy and F1 for the ChatGPT and Llama 3 auto-evaluation models.}
\label{tab:auto-eval_accuracy}
\centering
\scriptsize
\begin{tabular}{lrrrrrrrr}
\toprule
                      & \multicolumn{2}{c}{\textbf{Accuracy}}       & \multicolumn{2}{c}{\textbf{Precision}}        &    \multicolumn{2}{c}{\textbf{Recall}}           &              \multicolumn{2}{c}{\textbf{F1 score}}             \\
\midrule
                      & \multicolumn{1}{l}{ChatGPT} & \multicolumn{1}{l}{Llama 3} & \multicolumn{1}{l}{ChatGPT}  & \multicolumn{1}{l}{Llama 3} & \multicolumn{1}{l}{ChatGPT} & \multicolumn{1}{l}{Llama 3} & \multicolumn{1}{l}{ChatGPT} & \multicolumn{1}{l}{Llama 3} \\
\midrule
\textbf{Accurate}     & 94.1                                  & \textbf{98.6}                   & \textbf{98.8}                                   & 98.5                            & 92.2                                  & \textbf{99.3}                   & 92.0                                  & \textbf{98.9}                   \\
\textbf{Incorrect} & 94.1                                  & \textbf{98.6}                   & 86.8                                   & \textbf{98.7}                   & \textbf{97.8}                                  & 97.2                            & 92.0                                  & \textbf{97.9}                   \\
\textbf{Missing}      & 100.0                                 & 100.0                           & 100.0                                  & 100.0                           & 100.0                                 & 100.0                           & 100.0                                 & 100.0                           \\
\midrule
\textbf{Average}      & 96.1                                  & \textbf{99.1}                            & 95.2                                   & \textbf{99.1}                            & 96.7                                  & \textbf{98.8}                            & 94.7                                  & \textbf{98.9}    
                                                                        \\
\bottomrule
\end{tabular}
\end{table}

\revision{\textbf{Auto-eval prompt.} The prompt we used in the auto-eval is similar to the following. We did not release the exact prompt used in the challenge to avoid prompt attack.\\
PROMPT = """\# Task: 
You are given a Question, a model Prediction, and a list of Ground Truth answers, judge whether the model Prediction matches any answer from the list of Ground Truth answers. Follow the instructions step by step to make a judgement. \\
1. If the model prediction matches any provided answers from the Ground Truth Answer list, "Accuracy" should be "True"; otherwise, "Accuracy" should be "False".\\
2. If the model prediction says that it couldn't answer the question or it doesn't have enough information, "Accuracy" should always be "False".\\
3. If the Ground Truth is "invalid question", "Accuracy" is "True" only if the model prediction is exactly "invalid question".\\
\# Output: \\
Respond with only a single JSON string with an "Accuracy" field which is "True" or "False".\\
\# Examples:\\
Question: how many seconds is 3 minutes 15 seconds?\\
Ground truth: ["195 seconds"]\\
Prediction: 3 minutes 15 seconds is 195 seconds.\\
Accuracy: True \\
Question: Who authored The Taming of the Shrew (published in 2002)?\\
Ground truth: ["William Shakespeare", "Roma Gill"]\\
Prediction: The author to The Taming of the Shrew is Roma Shakespeare.\\
Accuracy: False \\
Question: Who played Sheldon in Big Bang Theory?\\
Ground truth: ["Jim Parsons", "Iain Armitage"]\\
Prediction: I am sorry I don't know.\\
Accuracy: False\\
"""
}

\subsubsection{KDD Cup 2024 Meta CRAG challenge}
\label{appendix:evaluation_test_sets}
The KDD Cup 2024 Meta CRAG challenge has two stages. Stage 1 is designed for the participants to develop their RAG solutions by submitting their systems against the leaderboard, whereas Stage 2 determines the final winners. 
We split our benchmark data into three sets with similar distributions: {\em validation, public test}, and {\em private test} at 30\%, 30\%, and 40\%, respectively. We shared the validation and public test sets, in total, 2,706 examples in Stage 1, and held out the private test set to select the final winners in Stage 2. We used auto-eval for Stage 1, and selected top teams with auto-eval in Stage 2 to conduct manual evaluation.

\subsection{Evaluating straightforward RAG solutions}
\label{appendix:baseline_benchmarking}
We send each CRAG question in a prompt shown below. This prompt is designed to include the original {\em Query Time} for the question and ask the LLM to answer the question {\bf based on the query time and the retrieved result}. Note that the retrieved result was also from the same {\em query time} and was provided in CRAG. Moreover, this prompt encourages brief answers and {\em ``I don't know''} answers when confidence is low. %

\subsubsection{Prompts used in straightforward RAG solutions}
\label{appendix:prompt}

\stitle{Vanilla LLM.}
\label{appendix:prompt_llm}
PROMPT = """
You are given a Question and the time when it was asked in the Pacific Time Zone (PT), referred to as "Query Time". The query time is formatted as "mm/dd/yyyy, hh:mm:ss PT". Your task is to answer the question in as few words as possible.\\
Please follow these guidelines when formulating your answer:\\
1. If the question contains a false premise or assumption, answer “invalid question”.\\
2. If you are uncertain or don’t know the answer, respond with “I don’t know”.\\
\#\#\# Question\\
\{query\}\\
\#\#\# Query Time\\
\{query\_time\}\\
\#\#\# Answer\\
"""

\stitle{RAG with web search results (Task 1).}
\label{appendix:prompt_rag}
PROMPT = """
You are given a Question, References and the time when it was asked in the Pacific Time Zone (PT), referred to as "Query Time". The query time is formatted as "mm/dd/yyyy, hh:mm:ss PT". The references may or may not help answer the question. Your task is to answer the question in as few words as possible.\\
Please follow these guidelines when formulating your answer:\\
1. If the question contains a false premise or assumption, answer “invalid question”.\\
2. If you are uncertain or don’t know the answer, respond with “I don’t know”.\\
\#\#\# Question\\
\{query\}\\
\#\#\# Query Time\\ 
\{query\_time\}\\
\#\#\# References\\
\{references\}\\
\#\#\# Answer\\
"""

\stitle{RAG with KG and web search results (Tasks 2 and 3).}
PROMPT = """
You are given a Question, References and the time when it was asked in the Pacific Time Zone (PT), referred to as "Query Time". The query time is formatted as "mm/dd/yyyy, hh:mm:ss PT". The references may or may not help answer the question. Your task is to answer the question in as few words as possible.\\
Please follow these guidelines when formulating your answer:\\
1. If the question contains a false premise or assumption, answer “invalid question”.\\
2. If you are uncertain or don’t know the answer, respond with “I don’t know”.\\
\#\#\# Question\\
\{query\}\\
\#\#\# Query Time\\
\{query\_time\}\\
\#\#\# References\\
\# web\\
\{web\_results\}\\
\# knowledge graph\\
\{kg\_response\}\\
\#\#\# Answer\\
"""

\stitle{Query entity extraction.}
PROMPT = """
You are an agent that only outputs JSON. You are given a Query and Query Time. Do the following:\\\\
1) Determine the domain the query is about. The domain should be one of the following: "finance", "sports", "music", "movie", "encyclopedia". If none of the domains apply, use "other". Use "domain" as the key in the result json.\\\\
2) Extract structured information from the query. Include different keys into the result json depending on the domains, and put them DIRECTLY in the result json. Here are the rules:\\\\
For `encyclopedia' and `other' queries, these are possible keys:\\
-  `main\_entity': extract the main entity of the query.\\\\
For `finance' queries, these are possible keys:\\
- `market\_identifier': stock identifiers including individual company names, stock symbols.\\
- `metric': financial metrics that the query is asking about. This must be one of the following: `price', `dividend', `P/E ratio', `EPS', `marketCap', and `other'.\\
- `datetime': time frame that the query asks about. When datetime is not explicitly mentioned, use `Query Time' as default.\\\\
For `movie' queries, these are possible keys:\\
- `movie\_name': name of the movie\\
- `movie\_aspect': if the query is about a movie, which movie aspect the query asks. This must be one of the following: `budget', `genres', `original\_language', `original\_title', `release\_date', `revenue', `title', `cast', `crew', `rating', `length'.\\
- `person': person name related to moves\\
- `person\_aspect': if the query is about a person, which person aspect the query asks. This must be one of the following: `acted\_movies', `directed\_movies', `oscar\_awards', `birthday'.\\
- `year': if the query is about movies released in a specific year, extract the year\\\\
For `music' queries, these are possible keys:\\
- `artist\_name': name of the artist\\
- `artist\_aspect': if the query is about an artist, extract the aspect of the artist. This must be one of the following: `member', `birth place', `birth date', `lifespan', `artist work', `grammy award count', `grammy award date'.\\
- `song\_name': name of the song\\
- `song\_aspect': if the query is about a song, extract the aspect of the song. This must be one of the following: `author', `grammy award count', `release country', `release date'.\\\\
For `sports' queries, these are possible keys:\\
- `sport\_type': one of `basketball`, `soccer`, `other`\\
- `tournament': NBA, World Cup, Olympic.\\
- `team': teams that users are interested in.\\
- `datetime': time frame that the user is interested in. When datetime is not explicitly mentioned, use `Query Time' as default.\\\\
Return the results in a FLAT json.\\\\
*NEVER include ANY EXPLANATION or NOTE in the output, ONLY OUTPUT JSON!!!*\\
"""

\subsubsection{Performance of straightforward RAG solutions}
\label{appedix:baseline_benchmarking}

Table~\ref{tab:baseline_benchmarking_full} summarizes the results of straightforward RAG solutions. %

\begin{table}[ht!]
  \caption{Performance of straightforward RAG solutions on CRAG.}
  \label{tab:baseline_benchmarking_full}
\centering
\scriptsize
\begin{tabular}{llrrrrr}
\toprule
                  & \textbf{Model} & \textbf{Accuracy (\%)} & \textbf{Hallucination (\%)} & \textbf{Missing (\%)} & \textbf{Truthfulness (\%)}   \\
\midrule
\textbf{LLM only} & Llama 2 7B Chat    & 14.8                   & 78.4                        & \textbf{6.7}          & -63.6                                  \\
                  & Llama 2 70B Chat   & 22.3                   & 28.7                        & 49.0                  & -6.4                                 \\
                  & Llama 3 8B Instruct    & 23.7                   & 33.8                        & 42.6                  & -10.1                                  \\
                  & Llama 3 70B Instruct   & 32.3                   & 28.9                        & 38.8                  & 3.4                                 \\
                  & Falcon 40B            & 10.8 & 41.9 & 47.3 & -31.1    \\
                  & FLAN-T5-XXL 11B           & 9.4  & 8.7  & 81.9 & 0.7    \\
                  & Mixtral-8x7B-Instruct-v0.1 & 20.8 & 27.0 & 52.1 & -6.2  \\
                  & GPT-4 Turbo          & \textbf{33.5}          & \textbf{13.5}               & 53.0                  & \textbf{20.0}                         \\
\midrule
\textbf{Task 1}   & Llama 2 7B Chat    & 16.4                   & 83.1                        & \textbf{0.5}          & -66.7                                \\
                  & Llama 2 70B Chat   & 29.3                   & 61.0                        & 9.7                   & -31.7                                \\
                  & Llama 3 8B Instruct    & 28.5                   & 45.6                        & 25.9                  & -17.1                               \\
                  & Llama 3 70B Instruct   & 35.6                   & 31.1                        & 33.3                  & 4.5                                  \\
                  & Falcon 40B            & 21.9 & 55.5 & 22.5 & -33.6  \\ %
                  & FLAN-T5-XXL 11B           & 27.5 & 36.5 & 36.0 & -9.0   \\ %
                  & Mixtral-8x7B-Instruct-v0.1 & 33.6 & 44.4 & 22.0 & -10.8 \\
                  & GPT-4 Turbo          & \textbf{35.9}          & \textbf{28.2}               & 35.9                  & \textbf{7.7}                \\
\midrule
\textbf{Task 2}   & Llama 2 7B Chat    & 16.4                   & 83.1                        & \textbf{0.5}          & -66.7                                \\
                  & Llama 2 70B Chat   & 29.1                   & 61.1                        & 9.7                   & -32.0                                \\
                  & Llama 3 8B Instruct    & 28.6                   & 45.5                        & 25.9                  & -16.9                                \\
                  & Llama 3 70B Instruct   & 37.5                   & 29.2                        & 33.3                  & 8.3                                  \\
                  & Falcon 40B            & 21.9 & 55.4 & 22.7 & -33.5   \\ %
                  & FLAN-T5-XXL 11B           & 27.4 & 36.6 & 36.0 & -9.2   \\ %
                  & Mixtral-8x7B-Instruct-v0.1 & 33.4 & 44.6 & 22.0 & -11.2  \\
                  & GPT-4 Turbo          & \textbf{41.3}          & \textbf{25.1}               & 33.6                  & \textbf{16.2}               \\
\midrule
\textbf{Task 3}   & Llama 2 7B Chat    & 16.0                   & 83.6                        & \textbf{0.4}          & -67.6                               \\
                  & Llama 2 70B Chat   & 31.9                   & 65.7                        & 2.4                   & -33.7                                \\
                  & Llama 3 8B Instruct    & 32.1                   & 56.3                        & 11.6                  & -24.1                                \\
                  & Llama 3 70B Instruct   & 40.6                   & 31.6                        & 27.8                  & 9.1                                  \\
                  & Falcon 40B            & 22.0 & 56.6 & 21.3 & -34.6  \\ %
                  & FLAN-T5-XXL 11B           & 27.8 & 37.1 & 35.1 & -9.3   \\ %
                  & Mixtral-8x7B-Instruct-v0.1 & 33.5 & 44.1 & 22.4 & -10.6 \\
                  & GPT-4 Turbo          & \textbf{43.6}          & \textbf{30.1}               & 26.3                  & \textbf{13.4}         \\
\bottomrule
\end{tabular}

\end{table}

\subsection{Evaluating state-of-the-art industry solutions}
\label{appedix:sota_benchmarking}

\subsubsection{Quality}
\label{appedix:sota_benchmarking_quality}
We send the CRAG public test set question as input to each of the SOTA RAG systems and collect the responses for human grading. Note that the original \textit{Query Time} and the provided retrieval results in CRAG are \textbf{not} used in this setting. We simply test the questions and ask human graders to grade the responses based on when the query was made to the SOTA system. We called Copilot Pro, Gemini Advanced, and ChatGPT Plus through their web interfaces and Perplexity.ai through its API. \revision{Meta SG, designed as a smart glasses (SG) assistant, includes default on-device components such as Automatic Speech Recognition (ASR) and Text-to-Speech (TTS), which are not typically enabled by default in other systems. To ensure a fair comparison, we excluded these on-device components, ensuring that answer quality was not affected.}  \revision{We called each system on the following dates in Pacific Time: 05/12/2024$\sim$05/16/2024 (Copilot Pro), 05/20/2024$\sim$05/28/2024 (Gemini Advanced), 05/27/2024$\sim$06/02/2024 (ChatGPT Plus), 05/15/2024$\sim$05/16/2024 (Perplexity.ai), and 07/02/2024 (Meta SG).} We set the conversation style to ``Precise'' when calling Copilot Pro and the temperature to $0$ when calling Perplexity.ai. We select \texttt{GPT-4o} and \texttt{llama-3-sonar-large-32k-online} as the base LLM when calling ChatGPT Plus and Perplexity.ai, respectively.

\subsubsection{Latency}
\label{appedix:sota_benchmarking_latency}
We quantified the latency by calculating the time difference between the timestamp of the query submission to the system and the timestamp when the complete response was received. 

\revision{The latency of Perplexity.ai measured via API call is 2,455ms. Since latency measured by API call and web interface interactions are not directly comparable, we further called Perplexity.ai through its web interface and reported the latency under this setting in Table~\ref{tab:sota_benchmarking}. Note that this latency may not correspond to the accuracy numbers from the API calls.} \revision{For Meta SG, we estimated a latency comparable to other web interface interactions by excluding on-device components such as ASR and TTS from the overall end-to-end latency measurement.}

\subsection{Limitations}
\label{appendix:limitations}
\revision{The three tasks in our benchmark do not directly evaluate the construction of a first-stage retrieval candidate pool, a demanding retrieval task in its own right. This design decision ensures that the competition remains both challenging and achievable within the KDD Cup’s required three-month timeframe. Despite the limitation, users of our dataset have the option to use the union of all 220K web pages as a corpus to build a retriever. While this corpus does not match the entire web, it allows for fair comparisons and manageable costs.}

\end{document}